\documentclass[conference]{IEEEtran}
\usepackage{times}

\usepackage[numbers]{natbib}
\usepackage{multicol}
\usepackage[bookmarks=true]{hyperref}
\usepackage{graphicx}
\usepackage{blindtext}
\usepackage{textcomp}
\usepackage[table,xcdraw]{xcolor}
\usepackage[mathlines, switch]{lineno}
\usepackage{amsmath,amsfonts,amssymb,amsthm}
\usepackage{mathtools}


\usepackage{xcolor}
\newcommand{\cmt}[1]{}

\long\def\ignorethis#1{}

\newcommand{\vc}[1]{\ensuremath{\mathbf{#1}}}

\DeclarePairedDelimiter\norm{\lVert}{\rVert}

\newcommand{\pctab}{\hspace{0.2in}}

\begin{document}

\title{Human Motion Control of Quadrupedal Robots\\ using Deep Reinforcement Learning}


\author{
    \IEEEauthorblockN{Sunwoo Kim\IEEEauthorrefmark{1}, Maks Sorokin\IEEEauthorrefmark{2}, 
    Jehee Lee\IEEEauthorrefmark{1}, 
    Sehoon Ha\IEEEauthorrefmark{2}}
    \IEEEauthorblockA{\IEEEauthorrefmark{1}Seoul National University, \hspace{3mm}
    \IEEEauthorrefmark{2}Georgia Institute of Technology}
    \IEEEauthorblockA{Email: sunwoo@mrl.snu.ac.kr, 
    maks@gatech.edu, jehee@mrl.snu.ac.kr, sehoonha@gatech.edu}
}

\maketitle

\begin{abstract}
A motion-based control interface promises flexible robot operations in dangerous environments by combining user intuitions with the robot's motor capabilities. However, designing a motion interface for non-humanoid robots, such as quadrupeds or hexapods, is not straightforward because different dynamics and control strategies govern their movements. We propose a novel motion control system that allows a human user to operate various motor tasks seamlessly on a quadrupedal robot. We first retarget the captured human motion into the corresponding robot motion with proper semantics using supervised learning and post-processing techniques. Then we apply the motion imitation learning with curriculum learning to develop a control policy that can track the given retargeted reference. We further improve the performance of both motion retargeting and motion imitation by training a set of experts. As we demonstrate, a user can execute various motor tasks using our system, including standing, sitting, tilting, manipulating, walking, and turning, on simulated and real quadrupeds. We also conduct a set of studies to analyze the performance gain induced by each component.(Video\footnote{Supplementary Video: \href{https://sites.google.com/view/humanconquad}{https://sites.google.com/view/humanconquad}})
\end{abstract} 
\IEEEpeerreviewmaketitle

\section{Introduction}
A tireless and invulnerable robotic worker entering dangerous environments has long been a dream for roboticists. Many researchers have approached this goal by developing autonomous robotic systems from various perspectives, such as model-based control or learning algorithms. However, a fully autonomous agent may not work in unforseen scenarios, such as disasters, where information is lacking. This limitation motivates the need for a more flexible control system that can be applied to novel scenarios.

We propose a human motion control interface that allows users to control robots using intuitive motions. This approach has great potential to overcome completely novel scenarios by combining humans' intuition with robots' motor capabilities. Traditionally, this problem has been approached with a model-based algorithm, such as the work of \citet{Ramos_2019_Tel} that projects human centroidal dynamics to the robot's space. Instead, our key idea is to exploit the recent advances in motion imitation learning~\cite{peng_2018_rl, Peng_2020_rcrl, Li_2021_rcrl2} that achieve realistic motion control on simulated characters or robotic creatures. We investigate quadrupedal robots as the target platform inspired by the recent success~\cite{hwangbo_2019_rcrl,Peng_2020_rcrl,Lee_2020_rcrl,kumar_2021_rcrl}.

Our motion control system consists of two main components: the motion retargeting module and the imitation control policy. The motion retargeting module takes a live human motion as input and translates it into the corresponding robot motion with proper semantics and dynamics. Then, the imitation policy tracks the retargeted motion based on onboard sensor information. We develop the motion retargeting module in a supervised learning fashion while training the imitation policy with deep reinforcement learning.

\begin{figure}
\centering
\includegraphics[width=0.8\linewidth]{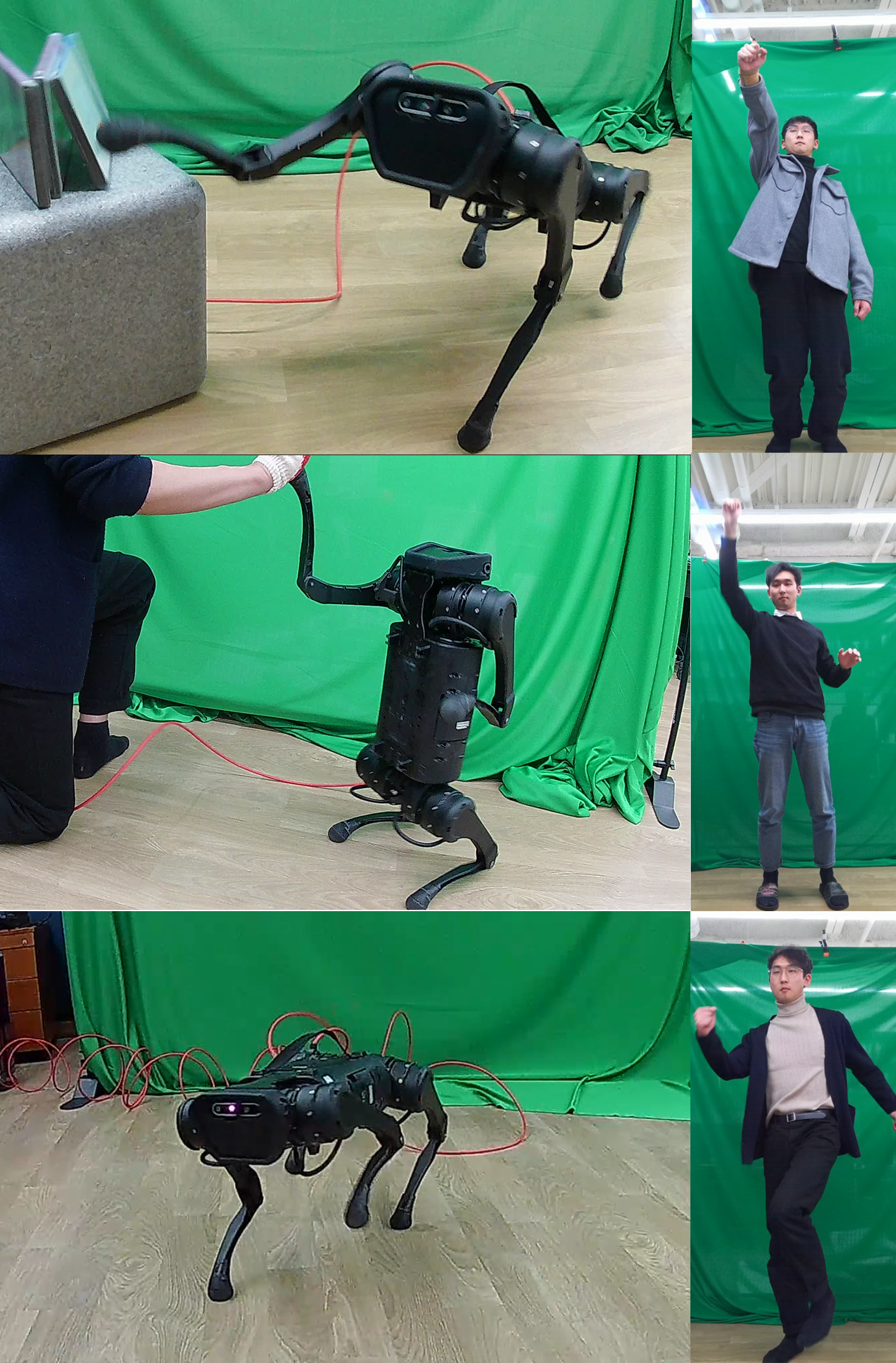}
\caption{We develop a novel control system that allows a user to control a quadrupedal robot on various tasks.
}
\end{figure}

However, we must address a few unique challenges to achieve the goal of developing a general motion control framework. First, we must deal with the ambiguous human motion that makes retargeting and control difficult. We mitigate this issue by adopting a hierarchical approach that learns a set of experts for both motion retargeting networks and control policies. We also develop a couple of post-processing techniques to improve the contact and temporal consistencies of the retargeted motion. Another key challenge is that our robot must imitate the target motion without accessing the future reference trajectory, often resulting in a conservative policy. We improve the training of the motion imitation policy by adopting curriculum learning, which gradually increases the difficulty over multiple tasks.

We demonstrate that our system allows a human user to execute various motor tasks with simulated and real quadrupedal robots using a consumer-grade motion capture system, Microsoft Kinect~\cite{microsoft_2018}. For instance, a user can control an A1 robot to approach the target and manipulate the object with both standing and sitting postures. A user can also tilt the robot's body to reach out to a distant object or avoid incoming obstacles. We evaluate our system by conducting an ablation study of essential components, including consistency corrections, curriculum learning, and domain randomization. We list our technical contributions as follows:

\begin{itemize}
    \item We design a novel human motion interface for a quadrupedal robot that requires minimal information about the task or the model.
    \item We develop an effective motion retargeting algorithm with contact and temporal consistency corrections.
    \item We improve the performance of motion imitation with curriculum learning and a hierarchical formulation.
    \item We demonstrate that a user can execute various motor tasks seamlessly on simulated and real robots.
\end{itemize}

\section{Related Work}
\subsection{Legged Robot Control}
\noindent \textbf{Legged robot control.} 
Striving for robust and dynamics robotic systems has enormously advanced the state of the hardware and software of the legged robots. By virtue of these advancements, legged robots can exhibit diverse, dynamic, and robust motion behaviors, allowing the robots to traverse challenging terrains or exhibit highly agile movements. The development of the hardware has enabled quadrupedal robots to perform agile motor skills while maintaining high stability~\cite{raibert_2008_legged, hutter_2016_legged, Bledt_2018_legged}. On the other hand, the development of bipedal robots has focused on the robustness of locomotion~\cite{bischoff_2002_legged, xie_2018_legged, kim_2021_legged}. Traditionally, designing effective motion controllers involves a lot of manual engineering and domain expertise. On the contrary, mathematical approaches like trajectory optimization~\cite{Raibert_1984_legged, geyer_2003_legged} and model predictive control (MPC)~ \cite{Horvat_2017_legged, gehring_2013_legged,Carlo_2018_legged,Ding_2021_legged}, leverage the optimization techniques to generating robot motions while alleviating the human-powered efforts in controller design process. Such optimization methods have enabled legged robots to complete the challenging control tasks such as locomoting on a slippery floor~\cite{Jenelten_2019_legged, carius_2019_legged}, traversing the rough terrain~\cite{Focchi_2020_legged}, recovering from the slip~\cite{Focchi_2018_legged} and even keeping the balance on a large ball in a physics simulation~\cite{yang_2020_legged}. However, the complexity of the real-legged robot dynamics usually forces these algorithms to either operate with a simplified robot model or design a task-specific controller. Our algorithm, on the contrary, allows the robot to learn a wide range of tasks without any task-specific dynamics modeling.

\noindent \textbf{Learning-based control.} Reinforcement learning (RL) control of physically simulated characters has led to a great performance in sophisticated motor skills such as walking, jumping, cart-wheel, and skating ~\cite{peng_2018_rl, lee_2019_rl, park_2019_rl,yu_2019_rl}. However, while controllers behave well in idealized simulated environments, they often struggle when transferred to the real world, exhibiting infeasible motor-control behaviors due to the difference between simulation and real-world, which is often referred to as the \emph{reality gap}. Some approaches propose to address the reality gap with conventional optimization methods such as MPC, allowing the policy to adjust on the real-robot~\cite{Iscen_2018_rcrl, tan_2018_rcrl, li_2021_rcrl, Xie_2021_rcrl}. On the other hand, others have investigated methods that leverage real-world data, such as learning on real robots ~\cite{haarnoja_2018_rcrl, ha_2020_rcrl, smith_2021_rcrl}, identifying system parameters~\cite{hwangbo_2019_rcrl}, or adapting policy behaviors~\cite{Peng_2020_rcrl, yu2020learning, kumar_2021_rcrl}. Instead, we leverage a Domain Randomization (DR) technique~\cite{peng2018sim, yu2018policy,openai2019learning, vuong2019pick, cubuk2019autoaugment, ruiz2019learning, Li_2021_rcrl2}, which randomizes domain parameters like mass, friction or PD gain during training in simulation to obtain more robust control policies while training only in simulation.
\par
\par

\subsection{Motion Imitation} 
Data-driven motion controllers have been proven effective for generating a wide range of physically plausible motions by leveraging motion capture data. Although kinematic approaches can provide interactive motion control~\cite{holden2017phase,  Bergamin_2019_rl,starke2019Neural, holden2020Learned, starke2020Local, starke_2021_rl}, they cannot be directly transferred to real-world due to the lack of physical plausibility. On the other hand, physics-based motion trackers~\cite{liu_2017_rl, liu_2018_rl} allow us to obtain natural motions in simulation, but its control design requires additional manual efforts, such as feature selection and motion processing. The recent RL-based formulation \cite{peng_2018_rl} provides an automated pipeline for developing effective motion imitation control policies from simple reward descriptions, which is capable of learning various motions on simulated characters~\cite{heess2017emergence,won2017How, won2018Aerobatics, peng_2018_rl,clegg2018Learning, lee_2019_rl,Min2019SoftCon, park_2019_rl, yu_2019_rl, luo202CARL, lee_2021_rl}, or even on a real quadrupedal robot~\cite{Peng_2020_rcrl} with manual motion retargeting. We adopt the concept of imitation objective to gain both physically correct motion and interactive control.
\par

\subsection{Motion-based Control}
Human motion control allows for direct control of the robot body based on the human body motion. Motion control schemes can liberate the human operator from the means of the commonly used control mechanisms (e.g. joysticks, keyboards), and allows the operator to better convey his or her intents to the robot controller. In this reason, human posture-based control has been widely studied in the fields of computer animation and robotics \cite{safonova_2003_tel,suleiman_2008_tel,baran_2009_tel,Yamane_2010_tel, albrecht_2011_tel, seol_2013_tel,Zheng_2015_Tel,Whitney_2016_tel,Ramos_2017_tel, Ishiguro_2017_Tel, Koenemann_2017_Tel, Ramos_2019_Tel, kim_2020_tel, choi_2020_tel, choi_2021_tel}.

\noindent \textbf{Human motion control for humanoid robots.}
Humanoid robots that inherit many human body features are seemingly a suitable platform for mimicking human body motions. Application of such anthropomorphic creatures ranges from human interaction~\cite{Whitney_2016_tel} to housekeeping teleoperation~\cite{Bajracharya_2020_tel} or even hazardous disaster rescue~\cite{Ramos_2017_tel}. With further expansions to smaller-sized humanoid control via motion imitation~\cite{Koenemann_2017_Tel}, which highlights the challenges of projecting the human posture to a new morphology. A number of methods~\cite{safonova_2003_tel, suleiman_2008_tel, albrecht_2011_tel} have been proposed to address the differences in the morphology configuration space, such as link length, joint limits, and degrees of freedom. Aside from morphology differences, timing issues emerge when expanding the controller from simple posture mapping to more dynamic motions, such as maintaining a balance. \citet{Zheng_2015_Tel} propose to integrate a time-warping objective to obtain a smoother motion-to-motion mapping control. Balancing has been further addressed by several approaches, such as Linear Inverted Pendulum (LIP) model safety constraints~\cite{Ishiguro_2017_Tel} or the balance feedback to the human~\cite{Ramos_2019_Tel}. In addition, safety has also been another main challenge during teleoperation. For instance, \citet{choi_2021_tel} proposed a shared-latent embedding retargeting algorithm to avoid self-collisions. Arduengo et al.~\cite{arduengo_2019_tel} proposed a technique for adapting the dynamics of the end-effector to switch between stiffness and compliance to obtain better safety.
\par

\noindent \textbf{Motion retargeting to non-humanoid characters.} Although some non-human-like characters as animals or alphabet shape characters have a different configuration from human beings, it might be possible to convey the semantics of the human posture to the character's posture~\cite{baran_2009_tel, Yamane_2010_tel, seol_2013_tel} with proper motion retargeting. Researchers have proposed various pose-to-pose motion retargeting algorithms with probabilistic pose-to-pose mapping~\cite{Yamane_2010_tel}, using semantic deformation transfer as mapping~\cite{baran_2009_tel}, or a feature selecting method~\cite{seol_2013_tel}. However, these methods focused mainly on posture mapping, which is hard to be extended for robot control. On the other hand, Kim et al.~\cite{kim_2020_tel} proposed the embedding of cyclic motion on the shared latent space, which a user to control an ostrich character in a $2$D physics simulation~\cite{kim_2020_tel}. In this work, we propose a new control framework that allows a user to operate a quadrupedal robot with motions, which has a different morphology from a human. We achieve real-time motion control for a variety of tasks, including walking, tilting, manipulation, and sitting, with a minimal amount of information about each task.

\section{Overview}
\begin{figure}
\centering
\includegraphics[width=0.99\linewidth]{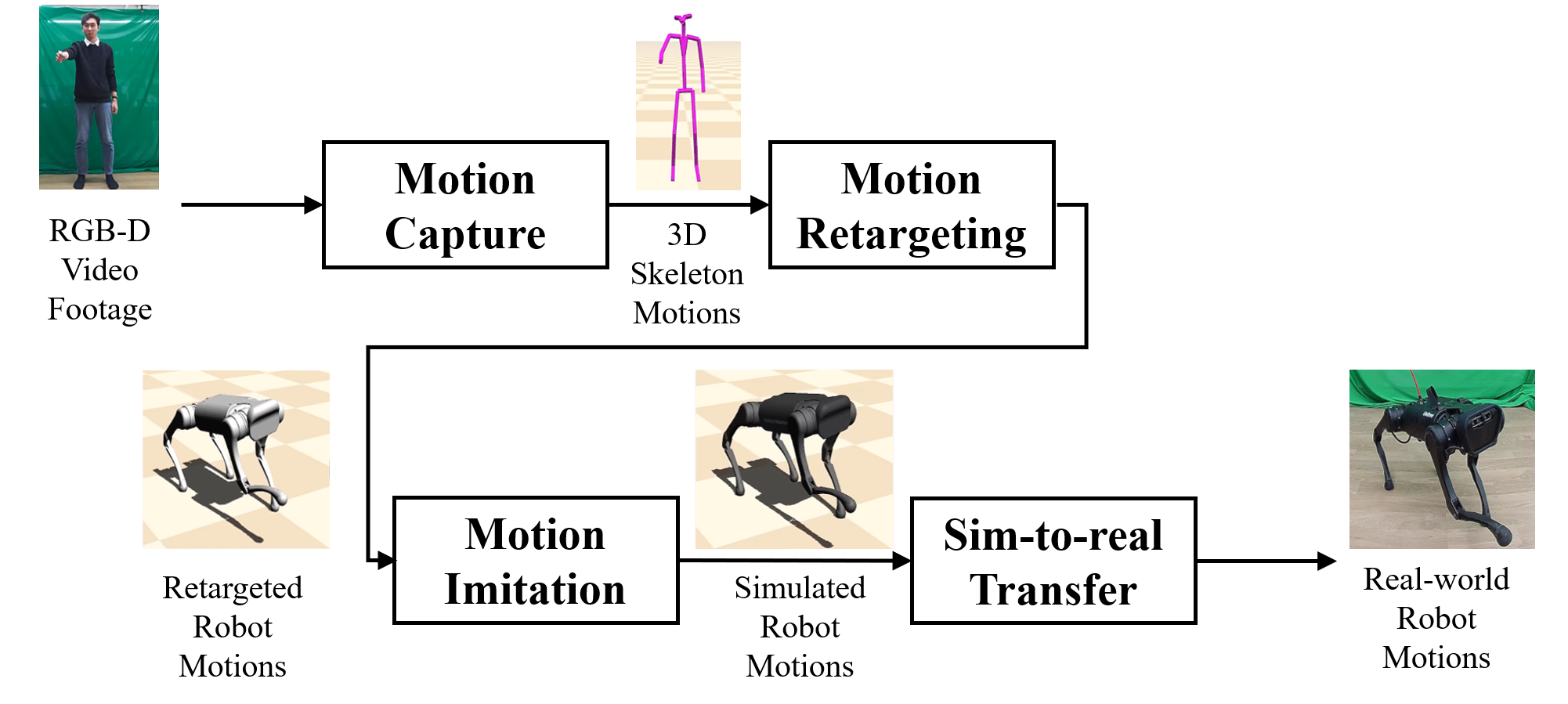}
\caption{Overview diagram. Our system takes a human motion as inputs and controls the robot via motion retargeting and motion imitation.}
\label{fig:overview}
\end{figure}

We develop a system for controlling a quadrupedal robot with a human operator's motions. Our system receives human motions from any motion capture system, which is Microsoft Azure Kinect~\cite{microsoft_2018} in our case. Then the motion retargeting module (Section~\ref{sec:retargeting}) converts the captured human into the corresponding robot reference motion that is physically valid and conveys proper semantics. To achieve this goal, we adopt a hierarchical approach of learning a set of experts mappers while applying optimization-based post-processing techniques. Then we learn a control policy that can imitate the given retargeted robot motion using deep reinforcement learning (Section~\ref{sec:imitation}). For more robust and flexible control, we develop robust expert policies using curriculum learning and combine them as a state machine with additional transition controllers. We illustrate the system overview in Figure~\ref{fig:overview}.

\section{Motion Retargeting} \label{sec:retargeting}
The first component of our motion-based control system is a motion retargeting module, which converts the user's motion into the corresponding robot motion. Many prior works have demonstrated successful human-to-humanoid motion mapping~\cite{safonova_2003_tel,suleiman_2008_tel, albrecht_2011_tel,Zheng_2015_Tel, Whitney_2016_tel, Koenemann_2017_Tel, Ramos_2017_tel, Ishiguro_2017_Tel,Ramos_2019_Tel, Bajracharya_2020_tel, choi_2021_tel}. However, our problem is unique in the sense that we have to find a mapping function between two very different morphologies without leveraging hand-engineered motion features, such as contact states or centroidal dynamics. Even worse, we have to address additional issues, such as the sparsity of the data and the required interactivity.

We tackle this problem by learning a set of expert networks and applying post-processing. Traditional techniques~\cite{dontcheva2003Layered, baran_2009_tel, Yamane_2010_tel} typically approach motion retargeting by solving optimization, but they tend to exhibit a slow turnaround time that is not suitable for interactive applications. And they also often require task-specific formulation~\cite{seol_2013_tel}, which makes the system hard to handle a wide variety of motions. On the other hand, learning-based approaches~\cite{choi_2021_tel} show impressive inference capabilities at interactive rates, but they are known to be data-hungry. In addition, they can also generate inconsistent or unexpected motions that can cause severe damage to the robot. We propose a motion retargeting algorithm that first infers the motion in a supervised learning fashion from a sparse dataset and corrects the inconsistency using optimization at the post-processing stage. The set of experts will be managed by an additional selector. Our framework allows us to build a fast and robust motion mapper that can be applied to various tasks.


\subsection{Motion Retargeting Network}

\begin{figure}
\centering
\includegraphics[width=0.99\linewidth]{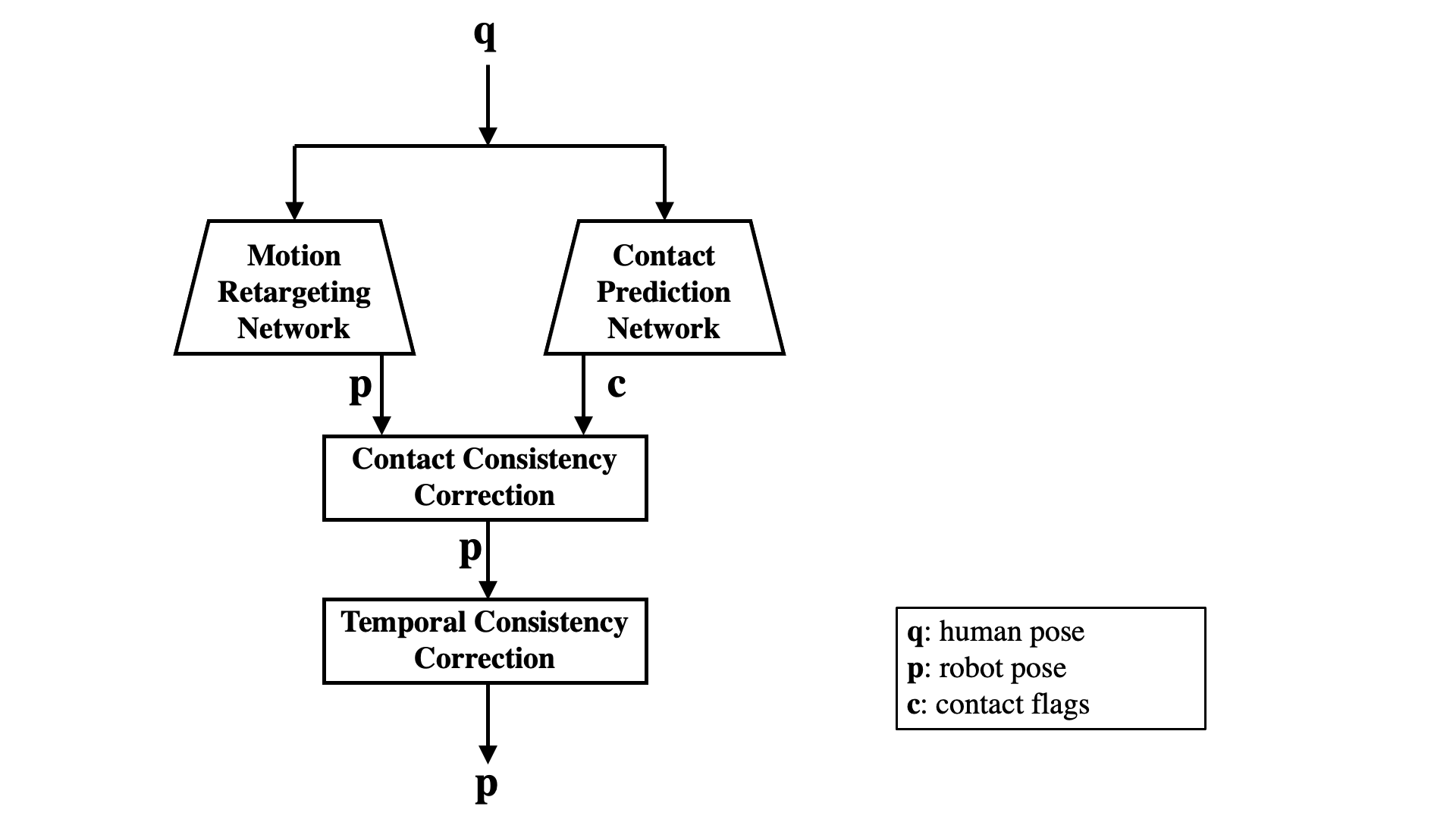}
\caption{The illustration of the motion retargeting module. The motion retargeting networks converts the given human motion $\vc{q}$ into the robot motion $\vc{p}$. The contact and temporal consistencies are maintained based on the inferred contact flags and previous history.}
\label{fig:mapper}
\end{figure}

In this section, we will explain how to learn a motion retargeting network for a single task. We aim to develop a motion mapper $f$ takes a human pose $\vc{q}$ as inputs and maps it into the corresponding robot pose $\vc{p}$. However, a simple pose-to-pose mapping can be ambiguous in periodic motions because a single pose does not contain any temporal information. For instance, we can interpret the same pose in an in-place marching motion as two different phases: \emph{swing up} or \emph{swing down}, which must be mapped to different quadruped poses. Therefore, our model learns to map a triplet of the human pose, velocity, and acceleration $(\vc{q}, \vc{\dot{q}}, \vc{\ddot{q}})$ to those of robots $(\vc{p}, \vc{\dot{p}}, \vc{\ddot{p}})$. We omit the derivatives in some figures and equations for brevity.


\noindent \textbf{Data preparation.}
We prepare the dataset $\mathcal{D}$ by collecting matching pairs of human and robot motions. First, we generate robot motions for sampled tasks. For example, tilting or manipulation tasks are synthesized by generating random goals and solving robot poses using inverse kinematics. Then we interpolate these key poses with a random time interval ranging from $1$ to $3$ seconds. 
For locomotion tasks, we generate a set of
walking motions with various gait parameters including body heights, foot clearance heights, and swing angles using a trajectory generator~\cite{Iscen_2018_rcrl}. Note that these motions can be reused for imitation policy training in Section~\ref{sec:imitation}.

Once we have the robot motions, we collect the matching human motion sequences. While showing robot motions, we ask a human user to act the ``corresponding motions'' based on the user's own intuition and record the motions using a motion capture system. 
We further manually process the motions to clean up noisy segments and fix asynchronous actions based on contact flags.
Once we obtain the motions, we compute the pose derivatives for both the user and the robot using finite differences with $\Delta t = 0.1$.

\noindent \textbf{Learning process.}
We use multi-layer perceptron (MLP) for training a mapper from the given dataset $\mathcal{D}$ (Figure~\ref{fig:mapper}). Our MLP consists of three leaky ReLU layers and one final hyperbolic tangent layer. Since a hyperbolic tangent function outputs the value between [-1, 1],  we shift and scale the outputs using the robot joint limit vector to finalize the joint angle of the robot. Our loss function is defined as follows:
\begin{equation}
	L_{map} = w_{ori} L_{ori} + w_{jnt} L_{jnt} + w_{dx} L_{dx} + w_{ddx} L_{ddx}.
\end{equation}
We omit the function arguments $\vc{p}, \vc{\dot{p}}$, $\vc{\ddot{p}}$, $\vc{\bar{p}}, \vc{\bar{\dot{p}}}$, and $\vc{\bar{\ddot{p}}}$ for brevity, where the former three are the outputs from the networks and the latter three are the target values.
The orientation loss $L_{ori} = d(\vc{p}^{root}, \vc{\bar{p}}^{root})$ compares the root orientation $\vc{p}^{root}$ in quaternion and its target value $\vc{\bar{p}}^{root}$ via a quaternion distance function $d$. The joint angle loss $L_{jnt}= \lVert \vc{p}^{jnt} - \vc{\bar{p}}^{jnt}\lVert^2$ is designed to match the joint angles $\vc{p}^{jnt}$ and their target values $\vc{\bar{p}}^{jnt}$. Two end-effector terms, $L_{dx} = \lVert\vc{\dot{x}} - \vc{\bar{\dot{x}}}\lVert^2$ and $L_{ddx} = \lVert\vc{\ddot{x}} - \vc{\bar{\ddot{x}}}\lVert^2$ compares the end effector velocities $\vc{\dot{x}}$ and accelerations $\vc{\ddot{x}}$ against their target values, $\vc{\bar{\dot{x}}}$ and $\vc{\bar{\ddot{x}}}$, respectively. Please note that these values can be derived from $\vc{\dot{p}}$ and $\vc{\ddot{p}}$. We set the weights $w_{ori}$, $ w_{jnt}$, $w_{dx}$, and $w_{ddx}$ as $0.3$, $1$, $0.001$, and $0.001$ for all the experiments, respectively. 

\subsection{Post-processing for Consistency}
The learned function often generates physically invalid motions in practice. This inconsistency slows the learning of a control policy and degrades the final motion's quality. To this end, we further clean up the motion at the post-processing stage to maintain contact and temporal consistency.

\noindent \textbf{Contact consistency correction.}
Contact consistency without foot skating is crucial to obtain physical plausibility. The violation of contact consistency destabilizes the robot balancing, hence leading to learning failure. To this end, we estimate four dimensional contact flags $\vc{c}_t$ and fix undesirable movements if they supposed to be in contact phases.

One possible approach to estimate $\vc{c}_t$ is to simply compare the current robot's foot heights against a certain threshold. However, we found that this approach yields undesirable discontinuous motions. Instead, we learn an auxiliary network that can predict smooth contact probabilities directly from human motions. We train this contact consistent network from the same training data using the following loss function:
\begin{equation}
	L_{cp}  = \lVert \vc{\bar{c}}_t (\vc{\bar{q}}, \vc{\bar{\dot{q}}})  -  \vc{c}_t(\vc{q}, \vc{\dot{q}}) \rVert^2,
\label{eq:contact_consistency_loss_function}
\end{equation}
where the contact probability $\vc{\bar{c}}_t$ is continuously estimated from the foot height and velocity: 1.0 if the height is $0$cm and the velocity is $0.0$cm/s, while 0.0 if the height is above $2$cm and the velocity is more than $60.0$cm/s. We apply inverse kinematics when the contact probability is greater than 0.5 to correct the contact feet to the previous frame's positions.


\noindent \textbf{Temporal consistency correction.}
We also invent an additional procedure to ensure the temporal consistency of the retargeted motions over multiple frames because abrupt movements on the real robot often cause dangerous situations. In our experiments, this was critical to deploy the proposed system to the real world by making the entire system more stable. To this end, we clip the joint angles with respect to the velocity limits, which is set to 120$^{\circ}/s$. 


\subsection{A Set of Experts for Multi-task Support}
\begin{figure}
\centering
\includegraphics[width=0.99\linewidth]{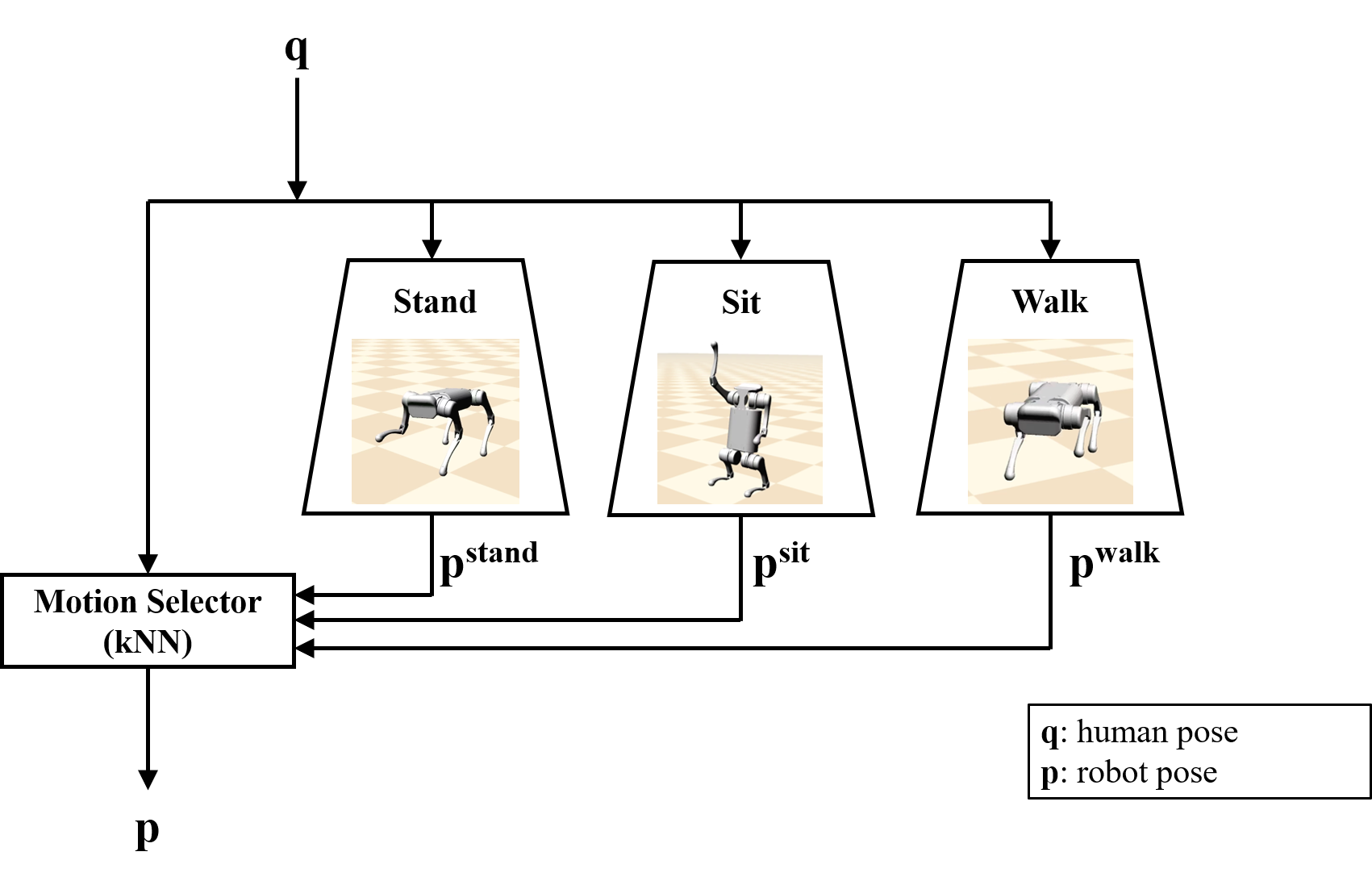}
\caption{We learn a set of expert motion retargeting networks for better accuracy in multi-task scenarios.}
\label{fig:ensemble}
\end{figure}
In our experiments, it becomes more difficult to obtain an accurate motion mapping when the human motions for multiple tasks are close to each other. To address this issue, we propose to use a hierarchical learning approach that manages a set of expert networks.\cite{yang_2020_rcrl, Won2020Scalable, Peng2019MCP, Frans2017Meta} We first learn three different motion retargeting networks for three robot states, \emph{stand}, \emph{walk}, and \emph{sit} (Figure~\ref{fig:ensemble}). Each network work can handle multiple tasks, such as \emph{manipulation-at-stand} or \emph{tilting-at-stand}. 
We query k-Nearest neighbors (kNN) over the input data to identify the expert associated with the closest data set; if one expert's training data is closer to the current one, we switch to the corresponding expert. We found that this results in an accurate mapping function while greatly reducing the time for manual engineering, such as hyperparameter tuning and data curation.

\section{Motion Imitation} 
\label{sec:imitation}
Once we generate the kinematic robot motions, the next step is to develop a control policy to imitate the given reference. We employ the motion imitation learning framework of \citet{peng_2018_rl} that allows natural and diverse motions in simulation, which has also been applied to a quadrupedal robot~\cite{Peng_2020_rcrl}.

Our problem is more ambitious than the others because we have to track a wide range of motions on real robots. In addition, our problem is facing additional unique challenges that make learning more difficult. First, we must imitate noisier references because they are from live human movements. Because a human cannot reproduce the exact same motion, our controller must be able to imitate similar motions with spatial and temporal noises. Second, our controller does not have access to the ``future'' reference motions. Indeed, this absence of future information often puts a robot into conservative states rather than actively tracking the references.

We aim to maximize the performance of motion imitation by introducing the following techniques. First, we design a hierarchical controller that learns three expert controllers for robot states, \emph{stand}, \emph{sit}, and \emph{walk}, while manually designing transition controllers between states. Second, we obtain effective expert controllers with curriculum learning, which is arranged over various difficulties and tasks. These inventions allow us to develop a practical controller that handles a wide range of motions on both simulated and real robots. 

\subsection{Background: Reinforcement Learning}
We formulate our problem as Partially Observable Markov Decision Processes (PoMDP) to utilize the reinforcement learning~\cite{sutton1999PG}. An each time step, an agent observes an observation $\vc{o}_t \sim \mathcal{O}(\vc{s}_t)$ emitted from the current state $\vc{s}_t$ and takes an action  $\vc{a}_{t} \sim \pi(\vc{a}_{t} | \vc{o}_{t})$ from its policy $\pi$. This results in the trajectory of the states and actions $\tau = \{( \vc{s}_{0},\vc{a}_{0}), (\vc{s}_{1},\vc{a}_{1}), \cdots (\vc{s}_{T},\vc{a}_{T}) \}$ where $T$ is the episode length. Our goal is to find the optimal policy that maximizes the expected return:
\begin{equation}
J(\pi) = E_{\tau \sim p(\tau|\pi)}[\sum_{t=0}^{T-1} \gamma^{t} r( \vc{s}_{t},\vc{a}_{t},  \vc{s}_{t+1})],
\label{eq:expected return}
\end{equation}
where $p(\tau|\pi)$ is a probability of the given trajectory $\tau$.

\subsection{Formulation of Motion Imitation}

We formulate the problem of imitating the given reference motion as PoMDP.

\noindent \textbf{Reference Motions.}
We take the generated robot trajectories that are used for training a mapping function in the previous section and use them as example reference motions for motion imitation learning. We injected a noise vector into reference motions to improve the robustness of the learned policy.

\noindent \textbf{Observation.}
The observation $\vc{o}_t = [\vc{z}_{t-3:t}, \vc{a}_{t-3: t-1}, \vc{\bar{p}}_{t-3:t}]$ consists of three components: robot sensor data, previous actions, and reference poses, with their corresponding histories. Each robot sensor data $\vc{z}_t$ is a $16$ dimensional vector from $12$ joint motor encoders and $4$ IMU orientation and angular velocity readings in pitch and roll axes. A history of previous actions $\vc{a}_{t-3: t-1}$ are also stored to make the problem more Markovian in the real world. The previous reference poses $\vc{\bar{p}}_{t-3:t}$ are also given to the robot. Please note that we do not have \emph{future} reference motions due to the nature of our problem, which makes the tracking task more difficult.

\noindent \textbf{Action.}
The action $\vc{a}_{t}$ defines as the PD target for the twelve joint motors of a robot. We apply the Butterworth low-pass filter with the cut-off frequency at $5$Hz to actions to generate smoother motions.

\noindent \textbf{Reward function.} 
The reward function encourages the agent to imitate the given reference motion while adapting to the physics simulation:
\begin{equation}
r_t = w^{main}r_{t}^{p} \cdot r_{t}^{e} \cdot r_{t}^{rp} \cdot r_{t}^{ro} \cdot r_{t}^{sp}  + w^{acc}r_{t}^{acc},
\label{eq:base reward function design}
\end{equation}
which adopts the multiplicative form inspired by previous works~\cite{lee_2019_rl, park_2019_rl}. The term $r_{t}^{p}$ refers to a joint imitation reward:
\begin{equation}
    r_{t}^{p} =  \exp{(s_{p}\sum_j \norm{\bar{\vc{p}}_{t}^{j} - \vc{p}_{t}^{j}}^2)},
\label{eq:pose reward function design}
\end{equation}
where $\bar{\vc{p}}$ and $\vc{p}$ are the target and current joint angles. The end-effector reward $r_{t}^{e}$ drives the robot to track the end-effector of the reference:
\begin{equation}
    r_{t}^{e} =  \exp{(s_{e}\sum_e \norm{\bar{\vc{x}}_{t}^{e}- \vc{x}_{t}^{e}}^2)},
\end{equation}
where $\bar{\vc{x}}_{t}^{e}$ and $\vc{x}_{t}^{e}$ are the target and current end effector positions. Similarly, the root position reward $r_{t}^{rp}$ and the root orientation reward $r_{t}^{ro}$ measures the differences in root position and orientation:
\begin{equation}
\begin{split}
    r_{t}^{rp} &=  \exp (s_{rp} \norm{\bar{\vc{x}}_{t}^\text{root}- \vc{x}_{t}^\text{root}}^2) \\ 
    r_{t}^{ro} &=  \exp (s_{ro} d(\vc{\bar{p}}^\text{root}_{t}, \vc{p}^\text{root}_{t})^2)
\end{split}
\end{equation}
by comparing the current root position $\vc{x}^\text{root}$ and orientation and $\vc{p}^\text{root}$ with respect to their target values, $\bar{\vc{x}}_{t}^\text{root}$ and $\vc{\bar{p}}^\text{root}_{t}$. Finally, we penalize the deviation from support polygon
\begin{equation}
    r_{t}^{sp} =  \exp{(s_{sp} d_{sp}(\vc{x}^\text{root}, \vc{p}^\text{root}, \vc{p})^2)},
\end{equation}
where $d_{sp}$ is the minimal distance to the support polygon. We only measure $d_{sp}$ when the robot is required to make at least three contacts: otherwise, $d_{sp}$ is defined as zero. Finally, we penalize excessive motions with the acceleration penalty term:
\begin{equation}
    r_{t}^{acc}=  \exp{(s_{acc}\sum_j \norm{\ddot{\vc{p}}_{t}}^2)}.
\end{equation}
For all experiments, we set weight terms $w^{main} = 0.9, w^{acc} = 0.1$ to emphasize the main mimicking term. The scaling coefficients are set to $s_{p} = 1.0$, $s_{e} = 20.0$, $s_{rp} = 20.0$, $s_{ro} = 5.0$ and $s_{sp} = 10.0$ respectively.

\noindent \textbf{Early termination.} The early termination accelerates the learning speed which is proven by many works~\cite{peng_2018_rl, peng2018SFV, won2018Aerobatics, Yu2018Learning}. We trigger the early termination when the trunk of the robot touches the ground and self penetrating contact happens. 

\noindent \textbf{Learning process.} We optimize policies with Proximal Policy Optimization~\cite{schulman2017proximal}. The policies are represented as feedforward networks that consists of two hidden layers with $256$ ReLu neurons. The PPO has a clipping range of $0.2$, learning rate of $0.00005$, the
discount factor is $\gamma = 0.95$, and the GAE parameter is $\lambda = 0.95$. The minibatch size is 128 for policy and value network. The max gradient norm is set to $0.5$.

\subsection{Curriculum Learning over Tasks and Difficulties}
While the above formulation works well for a single motion clip, our goal of learning a versatile policy for multiple tasks remains a challenging problem. In our experience, naive learning will result in a conservative policy that is stuck in a steady position to avoid falls while not trying to follow the target motion. To address this issue, we train an expert policy for the given state with a curriculum that expands the range of the motion and also expands the number of tasks.

For this purpose, we sort all the robot reference motions based on two criteria: a task type as a primary and difficulty of the task as a secondary. For instance, we train a control policy by training on the \emph{tilting-at-stand} task first and expanding the task set by adding the \emph{manipulation-at-stand} task. For both tasks, we gradually increase the difficulty of the task by measuring the range of reference motions.
Similarly, we train an \emph{walking} expert by expanding the curriculum from the \emph{walking forward} task to the \emph{turning left/right} task, with increasing turning rates.

\begin{table}
\centering
\resizebox{0.45\textwidth}{!}{%
\begin{tabular}{|c|c|c|}
\hline
\rowcolor[HTML]{C0C0C0} 
\textbf{Parameters}          & \textbf{Range}         & \textbf{Unit} \\ \hline
Link Mass                    & [0.75, 1.25] X default & kg           \\ \hline 
Ground Friction Coefficients & [0.5, 1.5]             & 1             \\ \hline
Proportional Gain                       & [0.7, 1.3] X default   & N/rad           \\ \hline
Derivative Gain                       & [0.7, 1.3] X default   & N$\cdot$s/rad       \\ \hline
Communication Delay          & [0, 0.016]             & sec           \\ \hline
Ground Slope                 & [0, 0.14]             & rad           \\ \hline
\end{tabular}%
}
\caption{Domain Randomization Parameters}
\label{tab:dr}
\end{table}

\subsection{Hierarchical Control with States}
Our motion includes multiple tasks, \emph{tilting}, \emph{manipulation}, and \emph{locomotion}, over three different robot states, \emph{stand}, \emph{sit}, and \emph{walk}, which yields different combinations such as \emph{tilting-at-stand} or \emph{manipulation-at-sit}. Instead of learning one monolithic policy, we learn three experts for three robot states and develop special transition controller that are called when the motion selector of the motion retargeting detects transitions. These transition controllers can be developed in multiple ways, such as model-based control or reinforcement learning, but we choose to reuse the existing motion imitation framework. The transition takes $1$ to $3$ seconds depending on the tasks and robot's state. During transitions, the robot executes the predefined policy while ignoring human motions. 

\subsection{Domain Randomization}
The gap between the dynamics of the simulation and the real world decreases the performance of policies trained in simulation that are deployed on a real physics system. We introduce Domain Randomization~\cite{peng2018sim}, that randomizes dynamics parameters during the training to obtain more robust control policies. The randomized dynamic parameters and their ranges are specified in Table~\ref{tab:dr}. We gradually increase the range of dynamic parameters with a curriculum similar to the method mentioned in subsection C. Detailed procedures are well mentioned in previous works~\cite{peng_2018_rl, Li_2021_rcrl2}.

\section{Results}
\begin{figure*}
  \centering
  \includegraphics[width=0.9 \textwidth]{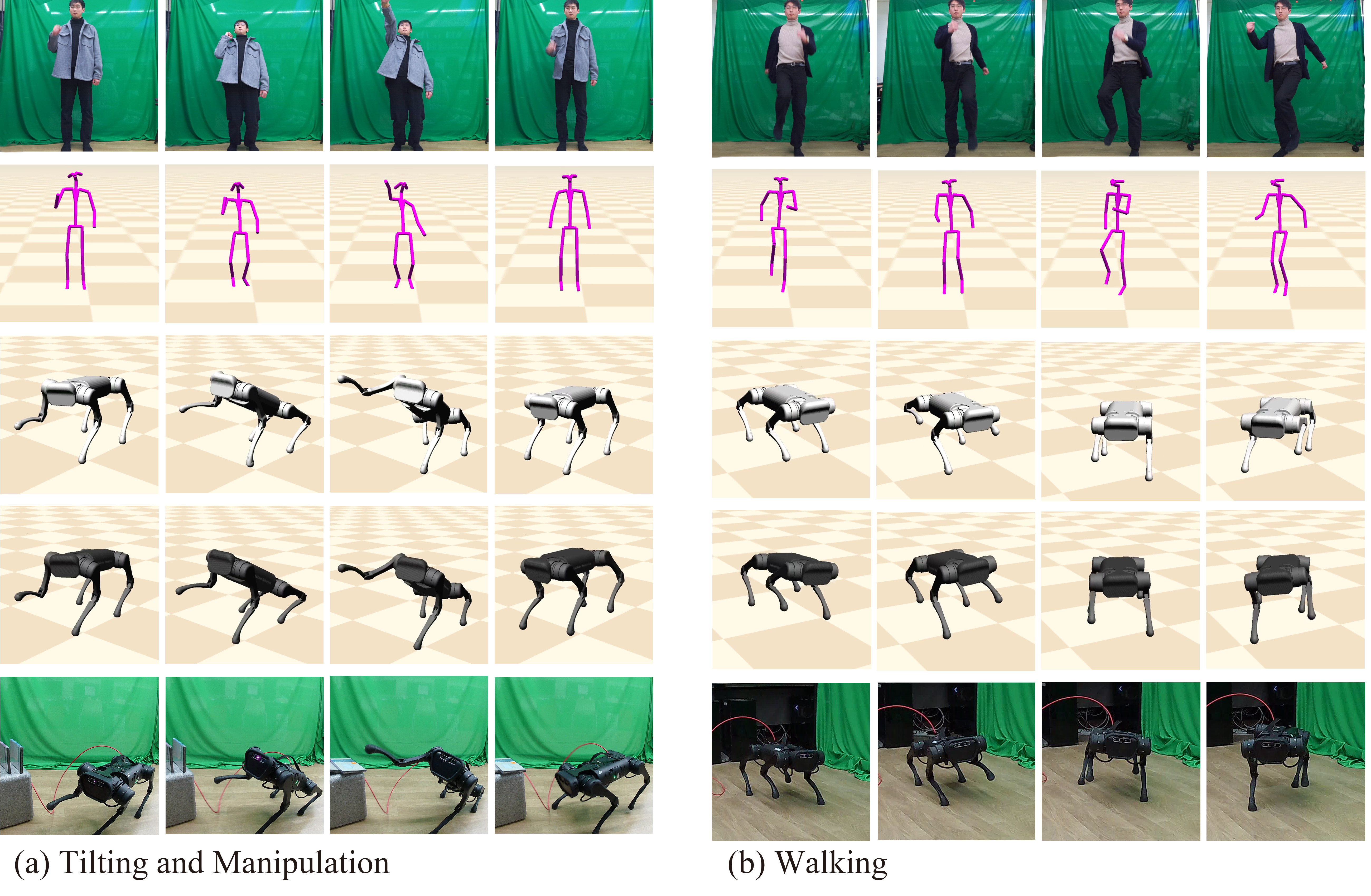}
  \caption{Motions of the (\textbf{a}) \emph{tilting and manipulation} and (\textbf{b}) \emph{walking} tasks. From the top row, we illustrate human video footage, human skeleton, retargeted robot motion, simulated motion, and real robot motion, at the corresponding time frames.}
  \label{fig:Demo}
\end{figure*}
\begin{figure*}
  \centering
  \includegraphics[width=0.9 \textwidth]{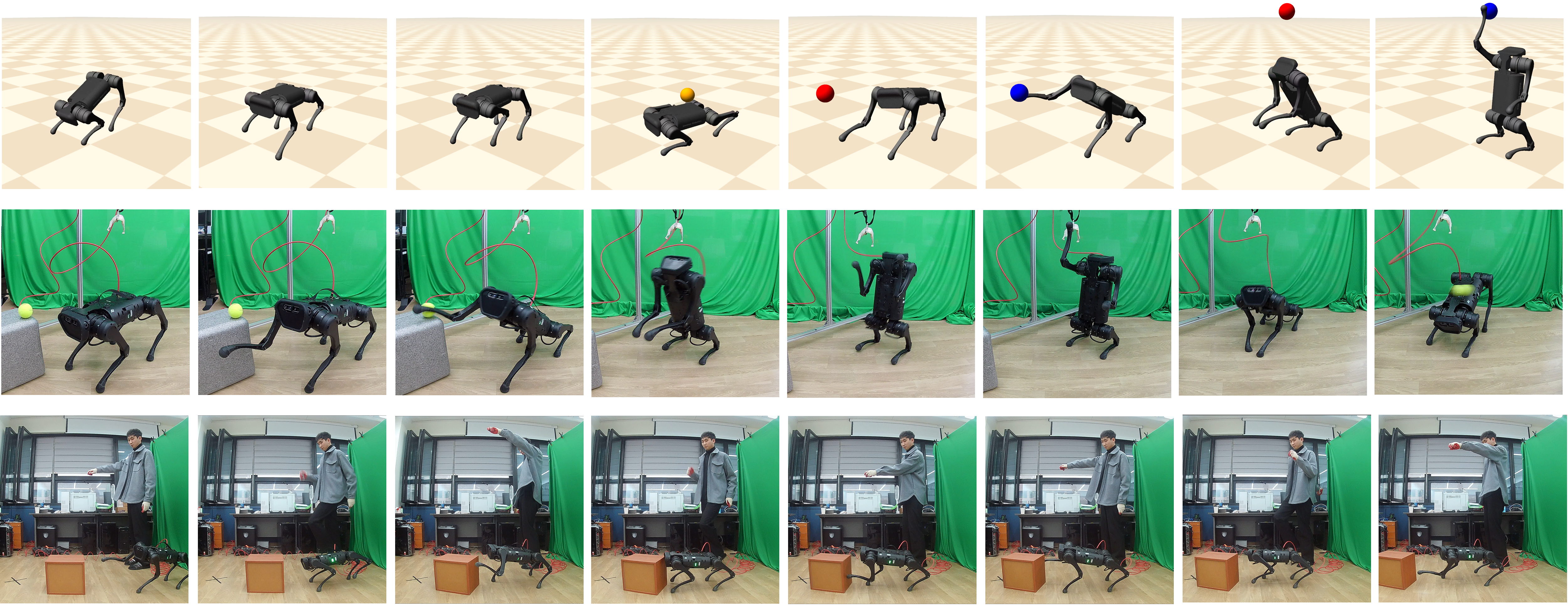}
  \caption{Composite task demos in simulation (\textbf{top}), real-world (\textbf{middle} (replay mode) and \textbf{bottom} (live mode)).}
  \label{fig:Demo2}
\end{figure*}

We design experiments to evaluate our framework from three perspectives. First, we evaluate the proposed system on a set of tasks in simulated and real environments. Second, we conduct an ablation study to evaluate the effectiveness of important components. Finally, we qualitatively compare our system with the previous motion-based interfaces.

\subsection{Experimental Setup}
We test our system on an A1 quadrupedal robot~\cite{a1}, which has three degrees of freedom for each leg and six under-actuated degrees of freedom for the root. We prepared $76$ to $522$ matching data pairs for training the mapper varying by the tasks. We train each expert policy using $1.2$ billion samples in the RaiSim~\cite{raisim} physics simulator. We conducted all the experiments with a desktop computer with Intel 16 core 3.60GHz i9-9900K CPU and GeForce RTX 2070 SUPER GPU. We capture a human motion using a Kinect~\cite{microsoft_2018}.

While all simulation demos are controlled interactively, we conduct real-world experiments in two modes: (1) the live mode that controls a real robot in an end-to-end fashion and (2) the replay mode that controls the robot to the prerecorded human motion trajectories. This is because fluctuating control delays could harm real robots. However, we do not feed the future trajectory information even in the replay mode, which is not available in the live mode. We annotate the experiment modes in the manuscript and supplemental videos.


\begin{figure}
\centering
\includegraphics[width=0.8\linewidth]{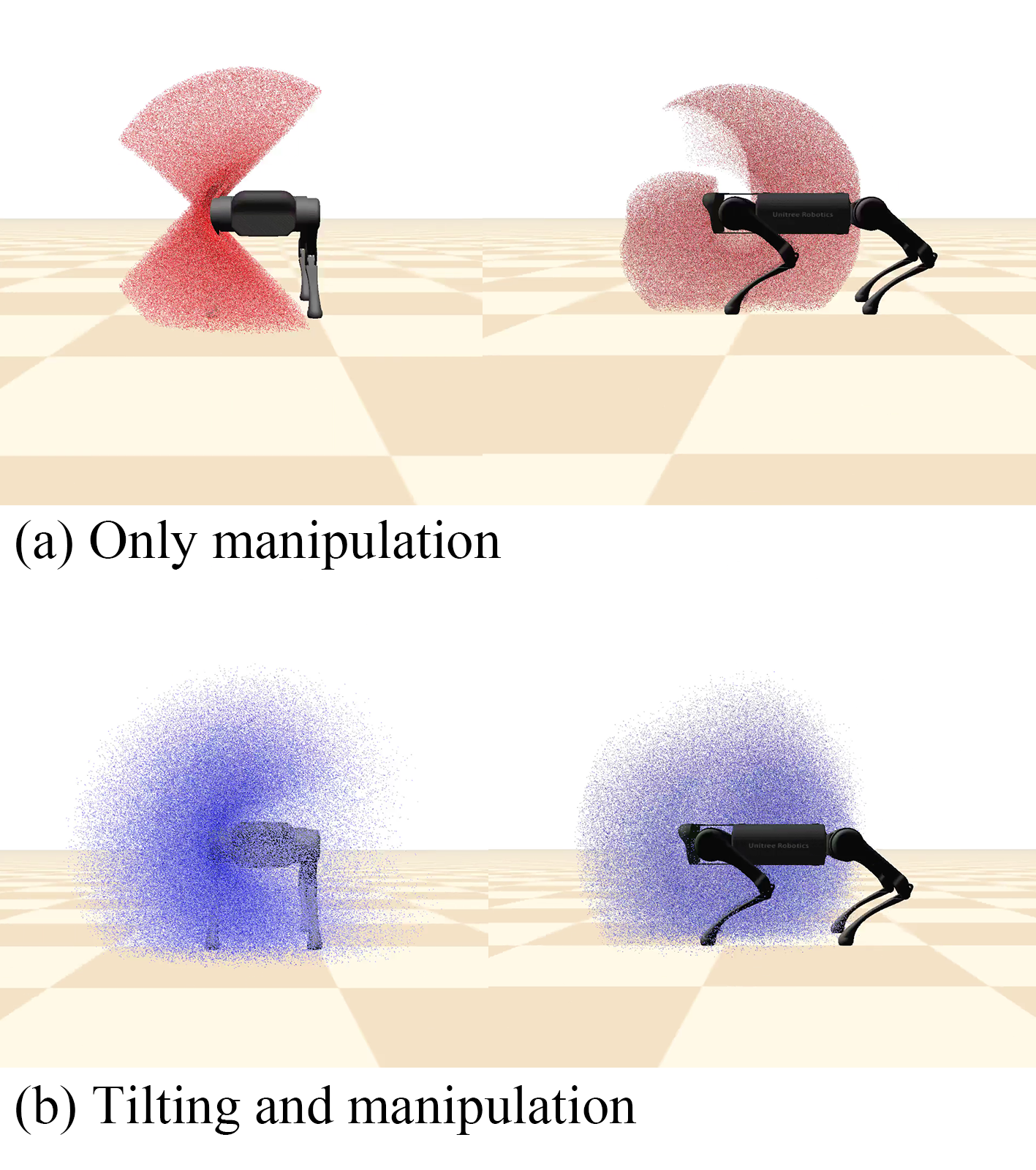}
\caption{Point clouds that illustrate the workspace of the right front leg when performing only manipulation(red) and manipulation with tilting(bottom) during \emph{standing}. The robot can reach approximately $2.7$ larger areas by simultaneously tilting its body.}
\label{fig:heatmap}
\end{figure}

\subsection{Motion Performance}
\noindent \textbf{Individual tasks.} 
Our system enables a user to control a diverse set of motor skills for A1 using human motions. In the \emph{stand} state, a robot can move its end-effector while simultaneously tilting its body. A robot can tilt its body $-40^{\circ}$ to $40^{\circ}$ for all $x,y,z$ axes, which is larger than the tilting range of the manufacturer's controller: $-20^{\circ}$ to $20^{\circ}$ for pitch and roll and  $-28^{\circ}$ to $28^{\circ}$ for yaw. Similarly, a robot can transit to the \emph{sit} state that allows the robot to use both arms and reach higher targets. During sitting, the robot can tilt $30^{\circ}$, $15^{\circ}$, and $7^{\circ}$ in pitch, roll, and yaw axes, respectively. Finally, a robot can walk at the speed of $0.0$m/s to $0.97$ m/s with the maximum turning rate of $15^{\circ}$ per second. The motion is less stable than a regular walking controller to prepare abrupt change of the speed at any moment. We illustrate the motions in Figure~\ref{fig:Demo} and the supplemental video. 

We also found that simultaneous manipulation and tilting provide a broader workspace, which is approximately $2.7$ times larger than manipulation without tilting. We compare the workspaces as point clouds in Figure~\ref{fig:heatmap}.

\par
\noindent \textbf{Composite tasks.}
Our system allows a user to switch between tasks seamlessly. In the simulation, we conduct an experiment with a sequence of the following tasks: (1) tilting the body to express a greeting, (2) walking forward to reach a target in $3$m, (3) dodging a thrown orange ball by crouching, (4) manipulating the target and (5) touching another target high in the air (Figure~\ref{fig:Demo2}~top). Similarly, we control a real robot to execute the following tasks: (1) hitting a tennis ball located at $0.42$m height, (2) touching a bone hanging high at $0.8$m height, and (3) dodging a thrown tennis ball (Figure~\ref{fig:Demo2}~middle). The robot must sit to achieve the second task because it cannot reach the bone while \emph{standing}. This long sequence is originally executed in simulation in the live mode and replayed on the hardware. These scenarios demonstrate the seamless transition capability of our control system. Finally, we control a real robot to push the box to the target position(X-mark on the floor) in real-time. This task is challenging because the box is located far from the initial position. To complete the task, we control the robot by repeating the following control tasks: (1) walk to the box and (2) push the box toward the target (Figure~\ref{fig:Demo2}~bottom).



\par
\noindent \textbf{Control responsiveness.} 
In real-time control, responsiveness is one of the most important criteria, which is even more critical for a quadrupedal robot with a floating base. Our control loop mainly consists of two stages: \emph{motion reconstruction} and \emph{control inference}. Control inference includes the main technical components, such as motion retargeting, post-processing, and querying the control policy, while motion reconstruction is inferring human $3$D poses from the Kinect. In our experience, the entire control inference takes less than $0.01$ second, which is fast enough for our $30$~Hz control loop. 
We stabilize the control frequency by skipping Kinect reading when the delay is significant and reusing the existing human motions from the previous frame.
\begin{figure}
\centering
\includegraphics[width=0.9\linewidth]{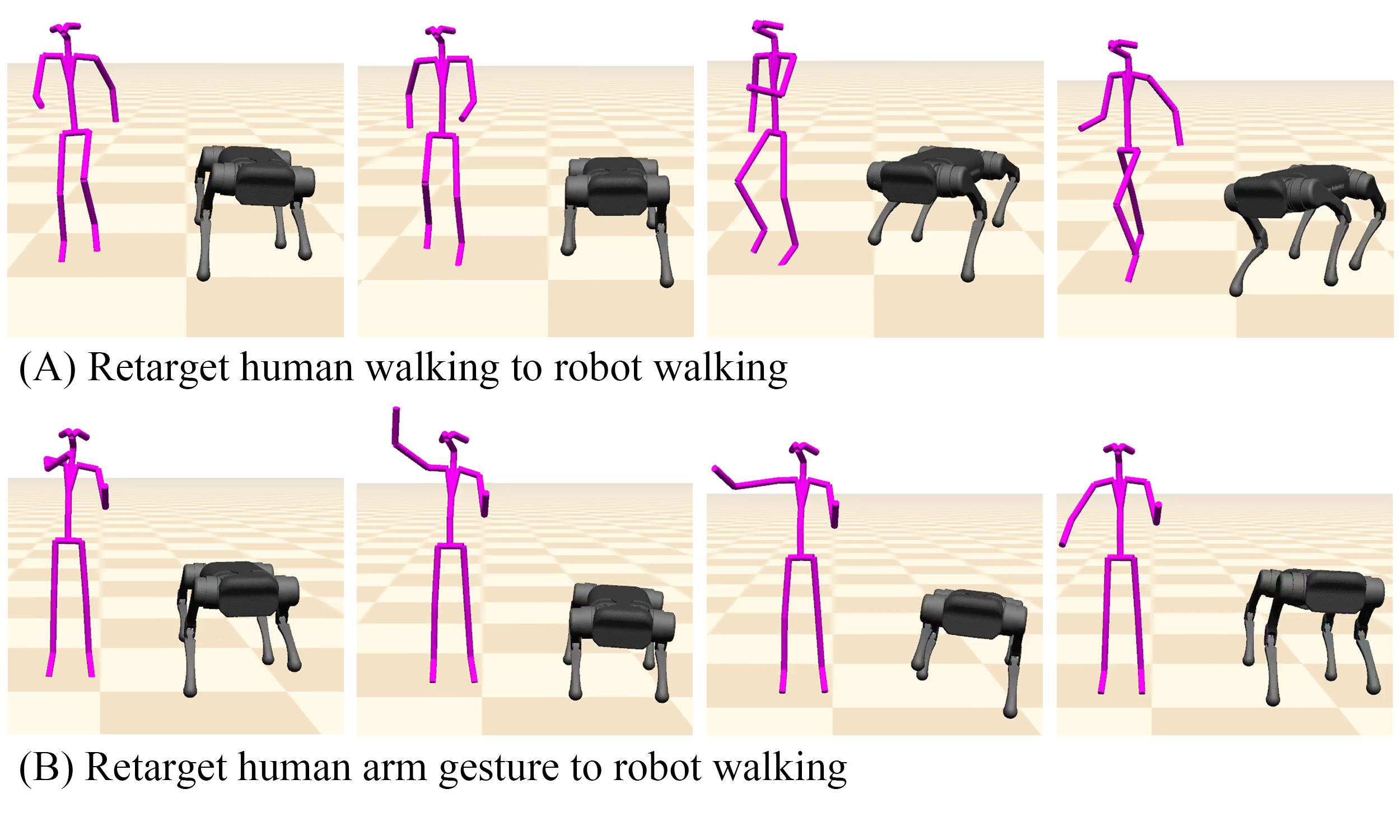}
\vspace{-0.2cm}
\caption{Different styles of mapping for \emph{walking}. The \textbf{top} shows a mapping with \emph{in-place marching} and the \textbf{bottom} shows a mapping with \emph{hand gestures}.}
\label{fig:Different style}
\end{figure}
\par
\noindent \textbf{Different mapping styles.}
Our system is flexible enough to support different styles of mapping for the same task. To demonstrate this, we generate two motion retargeting functions with (1) in-place marching motions and (2) circular hand gestures. Both mapping styles generate successful marching motions in the simulation (Figure~\ref{fig:Different style}).
\par
\noindent \textbf{Semantic mapping with manual features.}
We can also manually tune a mapping by selecting features for retargeting. For instance, we can build a mapping for the \emph{walking} task with explicit notions of the target velocities by extracting them from both human and robot motions. Although this explicit mapping offers slightly better motion quality, this requires domain-specific knowledge of the task.

\subsection{Analysis}
\begin{figure}
\centering
\includegraphics[width=0.8\linewidth]{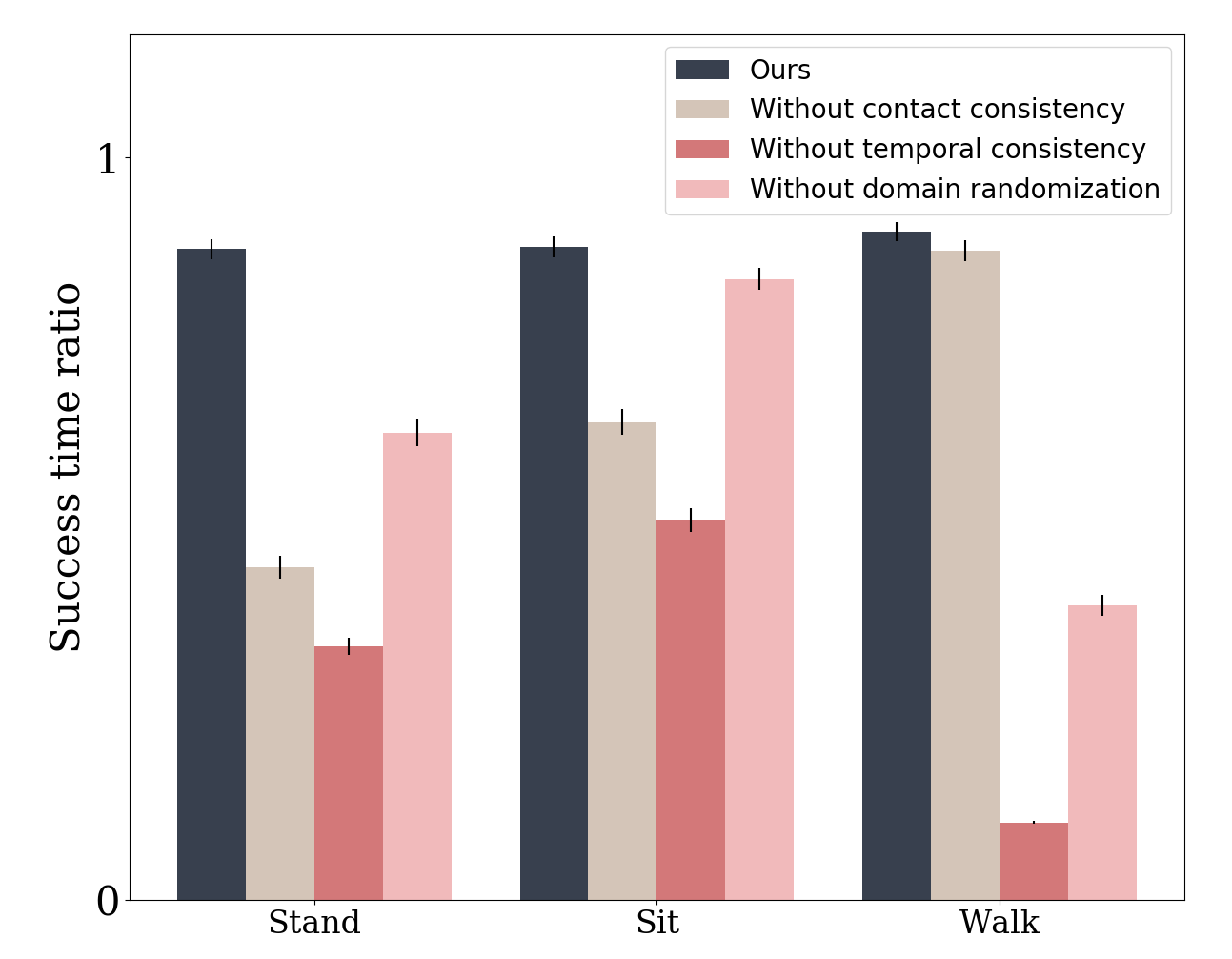}
\caption{Average success time ratio, which is the ratio of the termination time to the maximum episode duration. We conduct an ablation study with contact consistency, temporal consistency, and  domain randomization to evaluate their effectiveness.}
\label{fig:ablation}
\end{figure}

\noindent \textbf{Contact and temporal consistency.}
We evaluate the importance of \emph{contact consistency} and \emph{temporal consistency} corrections by conducting an ablation study. We generate $5120$ test episodes of tracking $10$ seconds of noisy trajectories perturbed from the ground truth robot motions. We compare methods based on the success time ratio, which is the ratio of the termination time to the maximum episode length. As illustrated in Figure~\ref{fig:ablation}, both components are essential for achieving the best performance. While contact consistency correction is more important for \emph{standing} and \emph{sitting} motions, temporal consistency seems crucial for \emph{walking} motions.

\begin{figure}
\centering
\includegraphics[width=0.9\linewidth]{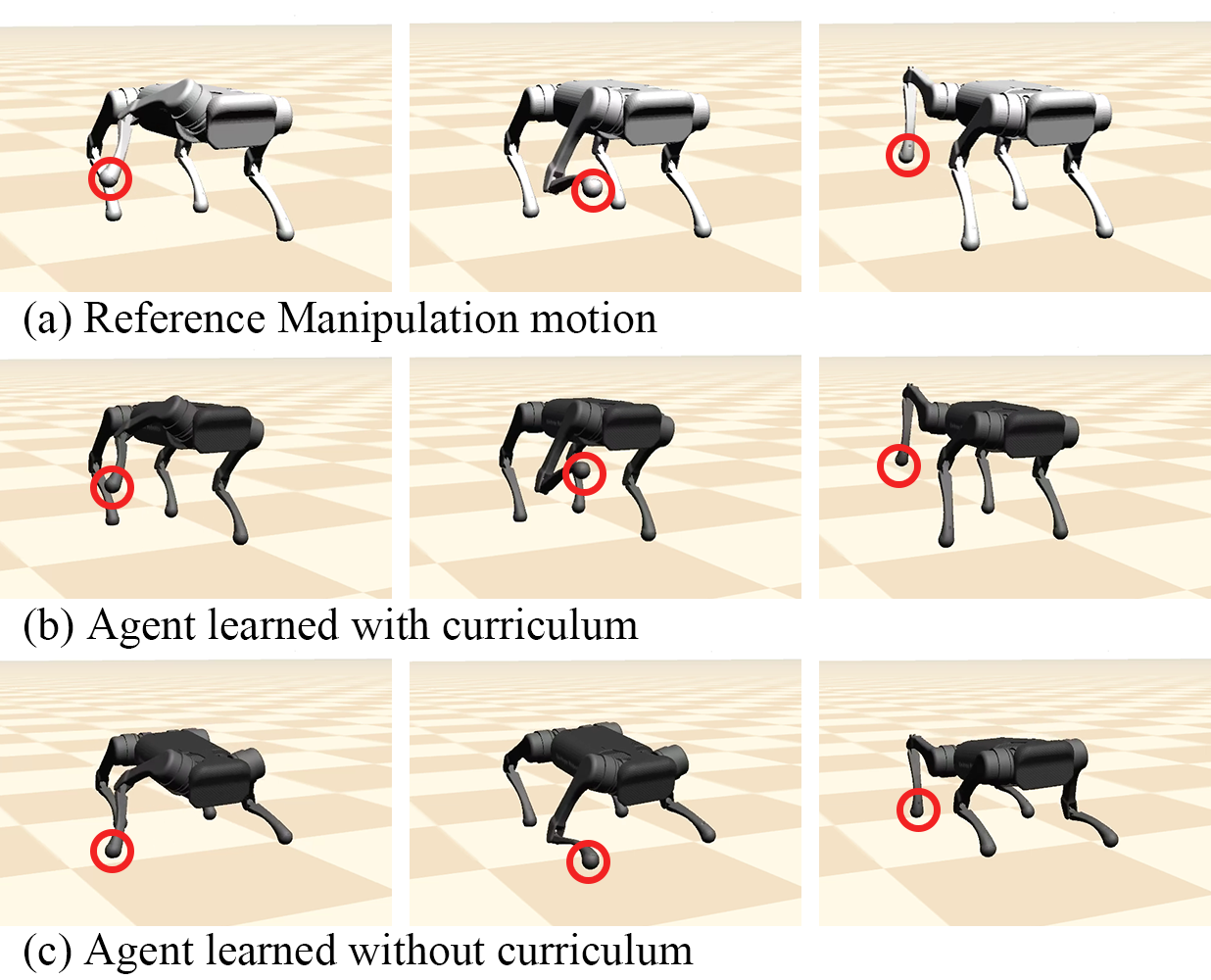}
\caption{Snapshots of the robot control to show the effectiveness of curriculum learning. The physically simulated agent tries to mimic the motion of reference (\textbf{Top}). While the policy trained with curriculum successfully mimic the reference (\textbf{Middle}), the policy without curriculum is stuck in local optimum (\textbf{Bottom}).}
\label{fig:curriculum ablation}
\end{figure}
\par
\noindent \textbf{Curriculum learning.}
In our experience, curriculum learning is essential for obtaining the best motion imitation performance. We do not evaluate the performance based on the success time ratio because policies without curriculum learning tend to survive until the last frame while showing conservative behaviors. Instead, we compare the quality of motions in Figure~\ref{fig:curriculum ablation} and the supplemental video. They illustrate the conservative behaviors of the policies without curriculum, which put all the feet on the ground and do not attempt to reach the target.
\par
\noindent \textbf{Domain adaptation.} Domain randomization (DR) has been one of the most effective techniques for overcoming the sim-to-real gap. We evaluate its effectiveness by measuring the success time ratio over $5120$ test cases with randomized dynamics. Figure~\ref{fig:ablation} shows that DR is essential for all the tasks. In addition, we conduct the sim-to-real experiment, where the policy without DR cannot complete the given motion: please refer to the supplemental video.

\noindent \textbf{Importance of future reference.} We often observe that the quality of the motion imitation is not as good as we expected. We hypothesize that the poor tracking performance is because our real-time motion tracking does not have information about the future reference trajectory. We verify this hypothesis by training an additional policy to track the fixed trajectory with future information and comparing the motion quality with the original agent. In our experience, the policy with future information generates more stable motions: please refer to the supplemental video for visual comparisons.

\subsection{Comparison to Other Methods}
\begin{table}
\centering
\resizebox{0.45\textwidth}{!}{%
\begin{tabular}{|l|c|c|c|c|c|}
\hline
\rowcolor[HTML]{C0C0C0} 
\textbf{Criteria} & \textbf{(A)} & \textbf{(B)} & \textbf{(C)} & \textbf{(D)} &\textbf{Ours} \\ \hline
\textbf{Mapping dimension}           & 2D
& 3D                                 & 3D                                 & 3D                                 & 3D                                 \\ \hline
\textbf{Real-Time}           & Y
& N                                 & N                                 & Y                                 & Y                                 \\  \hline
\textbf{Dynamics}            & Y
& N                                 & Y                                 & N                                 & Y                                 \\ \hline
\textbf{Mapping Flexibility} 
& N
& N                                 & N                                 & Y                                 & Y    \\ \hline
\textbf{Sim2Real}           
& N               & N                                 & N                                 & N                                 & Y                                 \\ \hline
\end{tabular}%
}
\caption{Comparison with previous human to non-humanoid control methods (A)\citet{kim_2020_tel}, (B)\citet{dontcheva2003Layered}, (C)\citet{Yamane_2010_tel} (D)\citet{seol_2013_tel}}
\label{tab:comparison}
\vspace{-0.3cm}
\end{table}

We compare our method with the previous human to non-humanoid character control approaches based on criteria that are meaningful for control in Table~\ref{tab:comparison}. \citet{kim_2020_tel}  showed the mapping that corresponds to the dynamical systems of two different morphologies, but it is limited to 2D cyclic motions. \citet{dontcheva2003Layered} proposed the concept of detecting the human gesture to search the matched motion pair of characters. This method only provides kinematic animation without the notion of dynamics. \citet{Yamane_2010_tel} successfully mapped human motions to non-humanoid characters with natural movements. However, this approach didn't aim to get real-time puppetry. \citet{seol_2013_tel} showed a flexible retargeting scheme that contains both agility and semantics but is limited to kinematic animations.
\par
Our framework has advantages over flexible mapping and real-time control compared to the other methods. Our framework supports various tasks without explicitly modeling task-specific dynamics due to the flexibility of motion retargeting and control schemes. We also achieve robust control by adopting the motion imitation learning with domain randomization.


\section{Conclusion And Future Work} 
\label{sec:conclusion}
We presented a human motion control system that allows a user to control quadrupedal robots using motion capture. The system has two main components: a motion retargeting module and a motion imitation policy. The motion retargeting module translates the captured human motion into robot motion with proper semantics through supervised learning and post-processing techniques. Then we train a control policy that can imitate the given retargeted motion using deep reinforcement learning. We further improve the control performance by leveraging a set of experts and curriculum learning. We evaluate the proposed motion control system on simulated and real-world quadrupedal robots by conducting various tasks, including standing, tilting, sitting, manipulating, walking, or their combinations.

Our work has a few limitations. First, we found that the Kinect's delay of $0.01$s to $0.06$s is significant for real-time control, particularly for a real robot, preventing us from conducting more real-world experiments in the live mode. While we mitigated this issue by training imitation policies with randomized control frequencies, it could not solve all the raised issues. Moreover, instability of the Kinect estimation system occasionally observes unexpected operator motions which often leads to control failure. We believe a motion capture system with more stability and higher rates, will be more suitable for real-time motion control applications. 

Another key observation is that the lack of future trajectory severely degrades the quality of motion imitation. We train policies to track versatile reference motions with frequent changes in target velocities and turning rates, often resulting in conservative policies with poor motion quality. One notable research direction will be to predict user intentions from history and leverage them to improve tracking performance.

We assume that the user and the robot are in the same space. However, we must release this assumption to achieve the goal of developing robotic workers in dangerous environments. We plan to combine the proposed system with virtual reality devices to provide more immersive experiences. This extension will raise new research questions, such as which robot sensory information is essential for users to operate the robot properly and how to deal with increased delays.

\section{Acknowledgements} 
\label{sec:Acknowledgements}
We would like to thank Phil Sik Chang and Visak Kumar for their contributions to this work.

\bibliographystyle{plainnat}
\bibliography{main}

\end{document}